# Predicting the clinical citation count of biomedical papers using multilayer perceptron neural network

Xin Li (*0000-0002-8169-6059*)

*School of Medicine and Health Management, Tongji Medical College, Huazhong University of Science and Technology, Wuhan 430030, Hubei, China*

Xuli Tang\* (*0000-0002-1656-3014;* ***Corresponding author***)

*School of Information Management, Central China Normal University, Wuhan 430079, Hubei, China*

Qikai Cheng (*0000-0003-3904-8901*)

School of Information Management, Wuhan University, Wuhan 430074, Hubei, China

**Abstract**

The number of clinical citations received from clinical guidelines or clinical trials has been considered as one of the most appropriate indicators for quantifying the clinical impact of biomedical papers. Therefore, the early prediction of the clinical citation count of biomedical papers is critical to scientific activities in biomedicine, such as research evaluation, resource allocation, and clinical translation. In this study, we designed a four-layer multilayer perceptron neural network (MPNN) model to predict the clinical citation count of biomedical papers in the future by using 9,822,620 biomedical papers published from 1985 to 2005. We extracted ninety-one paper features from three dimensions as the input of the model, including twenty-one features in the paper dimension, thirty-five in the reference dimension, and thirty-five in the citing paper dimension. In each dimension, the features can be classified into three categories, i.e., the citation-related features, the clinical translation-related features, and the topic-related features. Besides, in the paper dimension, we also considered the features that have previously been demonstrated to be related to the citation counts of research papers.  The results showed that the proposed MPNN model outperformed the other five baseline models, and the features in the reference dimension were the most important. In all the three dimensions, the citation-related and topic-related features were more important than the clinical translation-related features for the prediction. It also turned out that the features helpful in predicting the citation count of papers are not important for predicting the clinical citation count of biomedical papers. Furthermore, we explored the MPNN model based on different categories of biomedical papers. The results showed that the clinical translation-related features were more important for the prediction of clinical citation count of basic papers rather than those papers closer to clinical science. This study provided a novel dimension (i.e., the reference dimension) for the research community and could be applied to other related research tasks, such as the research assessment for translational programs. In addition, the findings in this study could be useful for biomedical authors (especially for those in basic science) to get more attention from clinical research.

# 1 Introduction

It is increasingly aware that the clinical impact of biomedical papers is of great significance to scientific activities in biomedicine, such as research evaluation, resource allocation, and clinical translation (Eriksson et al., 2020; Kryl et al., 2012; Thelwall & Maflahi, 2016; Zhang et al., 2018). The government, the funders, and the public are usually more concerned with biomedical papers' impact on the health practice than their academic impact (Annapureddy et al., 2020; Hutchins et al., 2019b). Papers with higher clinical citation counts can often provide useful evidence for clinical decision-making and have a higher potential to be clinically translated for health promotion (Hutchins et al., 2019b; Li & Tang, 2021). The clinical citation count that denotes the citations received from clinical guidelines or clinical trials has been considered as one of the most appropriate indicators for quantifying the clinical impact of biomedical papers in previous studies (Eriksson et al., 2020; Hutchins et al., 2019b; Kryl et al., 2012; Lewison & Sullivan, 2008; Thelwall & Kousha, 2016; Thelwall & Maflahi, 2016). However, the clinical citation count of biomedical papers consumes time to accumulate. Therefore, it is important to predict the clinical citation count of biomedical papers shortly after their publication.

Previously, citation count prediction of academic papers (CCPAP) has been widely studied with classification or regression models in the field of bibliometrics, and various factors have been demonstrated to be highly related to the citation counts of research papers, such as the writing style, the length of the abstract, and the research topic (Abrishami & Aliakbary, 2019; Huang et al., 2022; Li et al., 2015; Ma et al., 2021; Ruan et al., 2020). However, few studies have explored the clinical citation count prediction of biomedical papers (CCCPBP) since there was no comprehensive and reliable database that tracks citations among biomedical papers (Liang et al., 2021; Xu et al., 2020; Yu et al., 2021). We also don't know whether those features helpful in the CCPAP could be effective predictors for the CCCPBP. In fact, most approaches in previous studies on clinical citation analysis relied on domain experts and manual annotation, which are time-consuming and labor-intensive (Grant et al., 2000; Kryl et al., 2012; Lewison & Sullivan, 2008), making them unsuitable for analyzing large amounts of biomedical papers.

With recent advances in open science, a series of data sources with well-extracted metadata, including citation relationships among biomedical papers, have successively been released, for example, the Microsoft Academic Graph (Tang et al., 2020; Wang et al., 2020), the PubMed Knowledge Graph (Xu et al., 2020), and the iCite database (Hutchins et al., 2019a). Hutchins et al. (2019b) first trained a random forest classifier based on the iCite database to predict whether a biomedical paper will be cited by clinical guidelines or clinical trials in the future. They used the information on citation (2 years after publication) and different categories of Medical Subject Headings (MeSH), including Animal-related, Cell/Molecular-related, and Human-related MeSH of a biomedical paper, as the input features of the classifier and finally achieved an accuracy of 84% and an F1-score of 0.56. Furthermore, they argued that more research is needed to investigate how to early predict the future use of biomedical papers by clinical studies.

Hence, the purpose of the current study is to early predict the clinical citation count of biomedical papers in the future. We first selected 9,822,620 biomedical papers published between 1985 to 2005 from the PubMed Knowledge Graph (Xu et al., 2020). Then, we extracted ninety-one features for each paper from three dimensions, including the paper dimension, the reference dimension, and the citing paper dimension. For each dimension, the features can be divided into three categories, i.e., the citation-related features, the clinical translation-related features, the topic-related features. Meanwhile, in the paper dimension, we also included the "other features" that have previously been demonstrated to be highly related to the future citation count of academic papers, such as the length of the abstract, the readability of the abstract, the international collaboration, and the financial support. We used six machine learning algorithms and the ten-fold cross-validation method to train the regression models for predicting the clinical citation count of biomedical papers, and finally, a four-layer multilayer perception neural network model achieved the best performance.

## 2 Related work
### 2.1 Clinical citation of biomedical papers

Clinical citation has gained increasing recognition from the research funders as a potential indicator of the clinical impact of biomedical research (Kryl et al., 2012; Thelwall & Kousha, 2016; Thelwall & Maflahi, 2016). Theoretically, when a biomedical paper with financial support is clinically cited, we could assume that the money provided by the funders has been useful in clinical research and practice, suggesting the payback of the research investment (Boyack & Jordan, 2011). It is of continuing significance to evaluate the payback in biomedicine because of the large financial investment (Li et al., 2020).

One of the most authoritative sources for quantifying the clinical citation of biomedical research at the paper level is the clinical guideline (Eriksson et al., 2020; Grant et al., 2000; Kryl et al., 2012; Lewison & Sullivan, 2008; Thelwall & Maflahi, 2016), which are official documents generated by domain experts for guiding the prevention, diagnosis, and treatment of specific diseases. Being cited by clinical guidelines demonstrates that a biomedical paper may have a direct influence on health care (Yue et al., 2014). In addition, Thelwall & Maflahi (2016) investigated 327 clinical guidelines and their 6,128 references and found that citations from clinical guidelines could be a more accurate indicator of biomedical research's clinical value than academic citations or Mendeley statistics. However, it was difficult to systematically analyze clinical guides on a large scale because clinical guidelines in many countries did not list the papers they cite (Eriksson et al., 2020; Thelwall & Maflahi, 2016; Yue et al., 2014).

Another important source of counting the clinical citation of biomedical papers is the clinical trial. Influencing or entering into clinical trials is a significant way through which biomedical research can be finally translated into clinical practice. Therefore, being cited by clinical trials could also be an indication of the clinical impact of biomedical papers (Thelwall & Kousha, 2016; Zarin et al., 2011). Meanwhile, several classification models of biomedical papers were also proposed at different levels, such as the "research level" (Boyack et al., 2014; Lewison & Paraje, 2004; Narin et al., 1976; Narin & Rozek, 1988) at the journal level, the "triangle of biomedicine" (Weber, 2013) at the MeSH level, the Translational Science (TS) score (Kim et al., 2020) and the translational potential (Hutchins et al., 2019b) at the paper level. With these classification schemes, clinical papers can be identified from paper sets, and then the citations from clinical papers could be used for counting the clinical citations of biomedical papers.

In this paper, we count the clinical citation count of biomedical papers by using the citations from clinical guidelines and clinical trials. Different from Hutchins et al. (2019b), which detected whether a biomedical paper will be cited by a future clinical guidline or trial, we focus on how many clinical citations a biomedical paper will receive from clinical guidelines or trials in the future. It is believed that the more clinical citations a biomedical paper receives, the greater its clinical impact or value.

**2.2 Citation count prediction of academic papers (CCPAP)**
A brief review of the citation count prediction of academic papers (CCPAP) could offer insights into the clinical citation count prediction of biomedical papers (CCCPBP) because of the similarity of the two tasks. Nowadays, the most commonly used methods of citation count prediction are machine learning-based regressors, with which a series of important results have been yielded. For instance, Amjad et al. (2022) analyzed the author-related features and used a multiple linear regression model to select features for predicting citations of both journal and conference papers. They concluded that the first-year citations of the author and the total citations of the author were the two most useful features. Li et al. (2015) combined the support vector regressor (SVR) and the citation count trend of a paper to predict its citation count and achieved an $R^2$ of about 0.68. Jimenez et al. (2020) designed a set of features on the writing style of papers' abstracts and used a linear regression model to explore how these features predict the citation count of academic papers. Onodera & Yoshikane (2015) adopted a binomial multiple regressor to predict the citation count of academic papers and found that the features related to the references of papers, such as the number of references and the publication time of references, are more important than other features. Although the high similarity between the CCPAP and the CCCPBP, none of the previous studies have examined whether these useful factors for the CCPAP were also significant predictors for the CCCPBP.

With the re-emergence of the artificial intelligence, neural network-based methods such as the multilayer perceptron neural network with the Back Propagation (BP) algorithm (Ruan et al., 2020), the transformer (Huang et al., 2022), the Doc2vec and LSTM (Ma et al., 2021), the recurrent neural network (Abrishami & Aliakbary, 2019), and the convolutional neural network (Xu et al., 2019), have been increasingly used to predict citation count of academic papers and achieved acceptable performance. These neural network-based methods have strong generality and robustness, and they don't require features to be independent or the data to be normally distributed. However, as pointed out by Ruan et al. (2020), sequence learning models (such as the RNN and the LSTM) and convolutional neural network (CNN) models may not be suitable for the CCCPBP, whose features are not time series, word sequences or high-dimensionally visual data. Therefore, in this study, we finally selected the multilayer perceptron neural network (MPNN) model to predict the clinical citation count of biomedical papers. The MPNN algorithm has been demonstrated to have great advantages on feature learning for regression prediction with the backward propagation mechanism (LeCun et al., 1988; Tolstikhin et al., 2021). In addition, to confirm the effectiveness of our method, we also compared the MPNN model with the classical machine learning models and other neural networks, including the linear Regression (LR), the support vector regression (SVR), the k-nearest neighbor regression (KNNR), the random forest regression (RFR), and the eXtreme gradient boosting (XGBoost).

# 3 Methodology
## 3.1 Data and pre-processing

*3.1.1 Data collecting*

We collected the dataset for this study from the PubMed Knowledge Graph (PKG), which is a comprehensive, open access, and enhanced PubMed with rich and fine-grained information (Xu et al., 2020). The version of the used PKG is based on the PubMed 2021 baseline, which contains over 30 million papers, and it can be freely downloaded from its official website[1] and loaded into a local MySQL database for further analysis. For each paper, its bibliographic information (such as PMID, article type, title, keywords, MeSH terms, abstract, disambiguated authors, funding information, and author affiliations), well-extracted and normalized bio-entities (including disease, chemical/drugs, proteins/genes, species, and mutations), as well as its citation relationships with other PubMed papers are provided in the PKG. In addition, papers without titles or abstracts were excluded.

*3.1.2 Overall distribution of the clinical citation count of biomedical papers*

As shown in Fig.1, we first plotted the distribution of total and clinical citation count of biomedical papers. Fig.1a illustrates that the total citation counts of biomedical papers show a long-tailed distribution: most biomedical papers have no citations, and very few papers have more than 100, 000 citations. Comparing the distribution of clinical citations (Fig. 1b) with that of the total citation (Fig. 1a), we have several interesting findings. First, the two distributions are quite similar to the long-tailed patterns. This indicates that the prediction of clinical citation count of biomedical papers is possible because the prediction of citation count of papers has been widely explored in previous studies. Meanwhile, several differences between them can also be found. For example, the right figure is not dense as the left one, indicating that a considerable part of papers with citations were not clinically cited. Another example, none of the biomedical papers have more than 10,000 clinical citation counts. Besides, the paper frequency was reduced to 1 when the clinical citation count increased to 400.

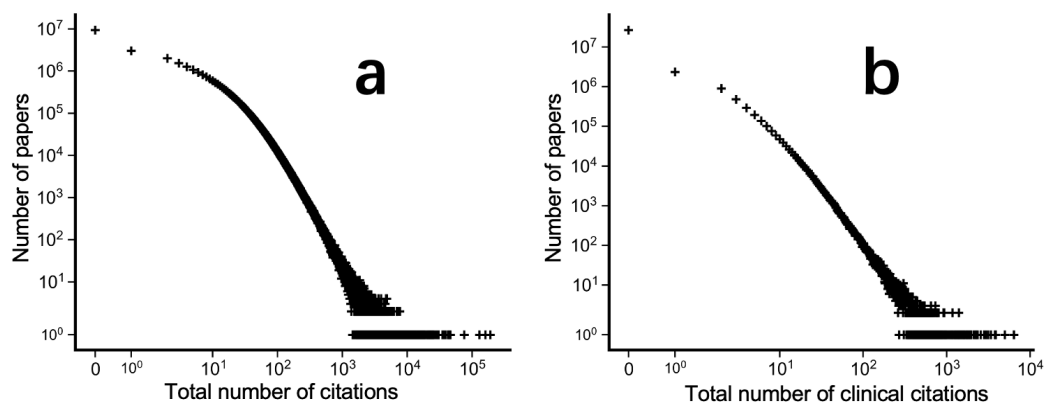

Fig 1. The distribution of citation counts of biomedical papers in PubMed, including (a) the total citation counts; and (b) the clinical citation counts.

3.1.3 Changes of the clinical citation count of biomedical papers over time

---

[1] http://er.tacc.utexas.edu/datasets/ped

We also analyzed the clinical citations of different categories of biomedical papers from the perspective of their ages. Specifically, we first defined the "age" of a paper is the years from its publication. For example, for a given paper (PMID: 10590187) published at the year 2000, then in the year 2000 and the year 2005, its ages are zero and five, respectively. We selected 532,395 biomedical papers published in the year 2000, and then classified these papers into three categories, including C, CA cand H according to the triangle of biomedicine proposed by Weber (2013). The C papers and CA papers represent papers within basic science, and the H papers represent papers closer to clinical science (Li and Tang, 2021). The statistical information about biomedical papers published in 2000 is shown in Table 1.

Table 1. The statistical information about biomedical papers published in 2000.

|  | ALL papers | C papers | CA papers | H papers |
| --- | --- | --- | --- | --- |
| Number | 532,395 | 27,547 | 68,057 | 197,239 |
| Proportion | 100% | 5.17% | 12.78% | 46.30% |

The cumulative distribution of the percentage of biomedical papers cited by clinical papers (i.e., the number of clinical citations is greater than zero) by the age of paper is shown in Fig.2a, which reveals three interesting findings. First, all the four curves first show a trend of rapid growth and then keep stable as the age increased. Second, the percentages of C or CA papers that were clinically cited is much lower than those of the overall papers (blue curve), while the percentage of H papers that were clinically cited is much higher than that of the overall papers. This finding indicates that a biomedical paper whose research content is closer to clinical science, is more likely to be clinically cited. Until 2020, 27% of biomedical papers published in 2020 have been clinically cited, including 38% of H papers, 13% of CA papers and 4% of C papers. Third, there is an obvious turning point when the age of paper reaches five. Then, the slope of each curve continues to decline. When the paper is fifteen-year-old, the curves tend to be stable. This illustrates that whether a biomedical paper is clinically cited or not, will tend to stabilize during five to fifteen years after its publication. Therefore, it is appropriate for us to set the citation window as fifteen years.

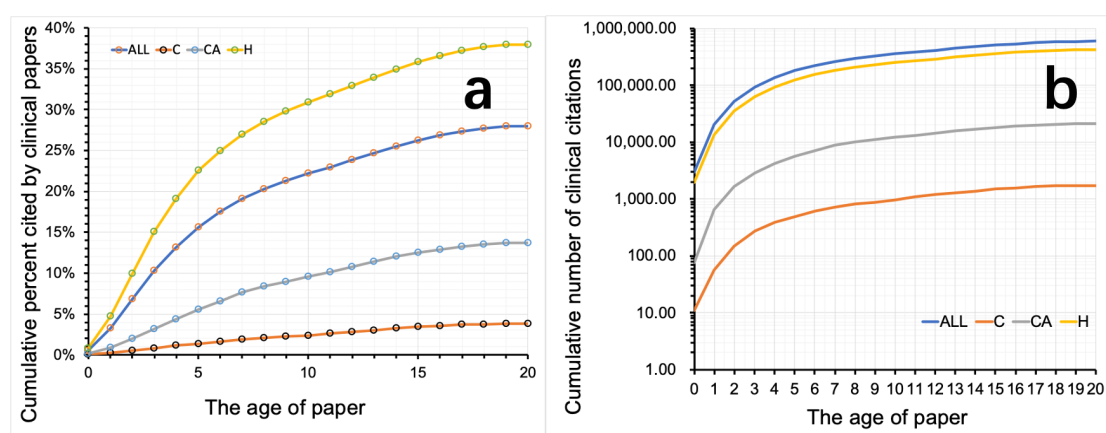

Fig 2. The clinical citations of biomedical papers published in the year 2000.

Fig. 2b displays the cumulative distribution of clinical citation count of biomedical papers published in 2000, from which we find that the cumulative clinical citation count shows an increase in the beginning and then keep stable as the age increases. Particularly, when the age of paper is larger than five, all the curves tend to be stable. This also indicates the clinical citation count of biomedical papers tend to be stable five year after their publication. Fig. 3 shows the distribution of the average clinical citation count of biomedical papers published in 2000 by the age of paper. From Fig. 3, we find that, for all ages, the average clinical citation count of H papers ranks the first place, followed by that of overall papers, and that of C or CA papers has been lower than overall papers. This also means that the paper whose research content is closer to clinical science is more likely to be clinically cited.

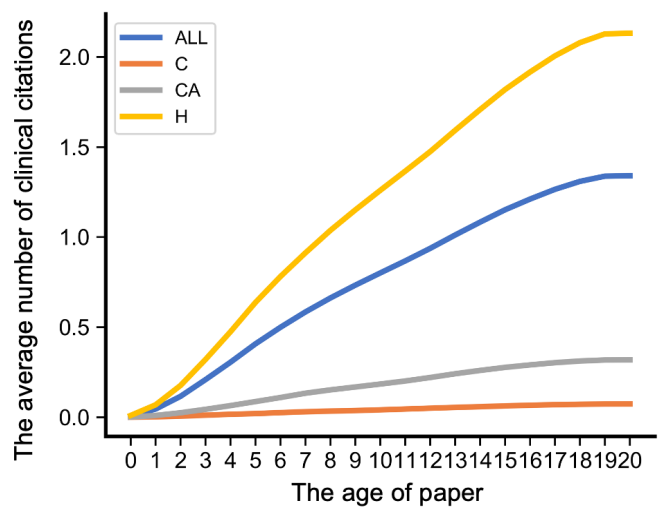

Fig 3. The distribution of the average clinical citation count of biomedical papers published in 2000 by the age of paper.

*3.1.4 Training and testing data*

Based on the above analysis, we finally selected PubMed papers published from 1985 to 2005 as our data source for prediction, because the clinical citation count of biomedical papers will remain stable fifteen years after their publication. There is a total of 9,822,620 biomedical papers published during these 21 years, and the number of clinical papers whose article types are clinical trials and clinical guidelines is 378,876. The average number of references for these 378,876 clinical papers is 26.7, i.e., the total number of these clinical papers is 10,115,989, which is larger than 9,822,620. In theory, every paper published during 1985-2005 may be clinically cited. In fact, only about 30% of these papers were clinically cited. This indicates that the dataset is of good quality for the task of CCCPBP. Finally, we randomly split the dataset into training data and testing data at a ratio of 4:1 (Huang et al., 2022; Ruan et al., 2020). The statistical information on the dataset is shown in Table 2.

Table 2 Statistical information on the dataset.

| Item | Value |
| --- | --- |
| Publication time (Year) | 1985-2005 |
| Total number of PubMed papers | 9,822,620 |
| Total number of clinical papers | 378,876 |
| Average number of references for the clinical papers | 26.7 |
| Number of PubMed papers in training data | 7,858,096 |
| Number of PubMed papers in testing data | 1,964,524 |
| The ratio between the number of training data and test data | 4:1 |

## 3.2 Feature extraction, calculation, and normalization

We treated the prediction of clinical citation count of biomedical papers as a regression task, such that our method could predict biomedical research with high potential for clinical translation in time. The prediction target that is a real number (i.e., $Y \in \mathbb{R}$) was defined as the number of the clinical citation count of a biomedical paper. Each biomedical paper was associated with ninety-one features, which can be classified into three dimensions, including the paper dimension, the reference dimension, and the citing paper dimension. As shown in Fig. 4, for a given paper A in purple that we are interested, the papers in red represent the set of all the references that were cited by the paper A, and the papers in green means the set of citing papers that cited the paper A. Each dimension consists of multiple categories of features, including the citation-related features, the clinical translation-related features, and the topic-related features.

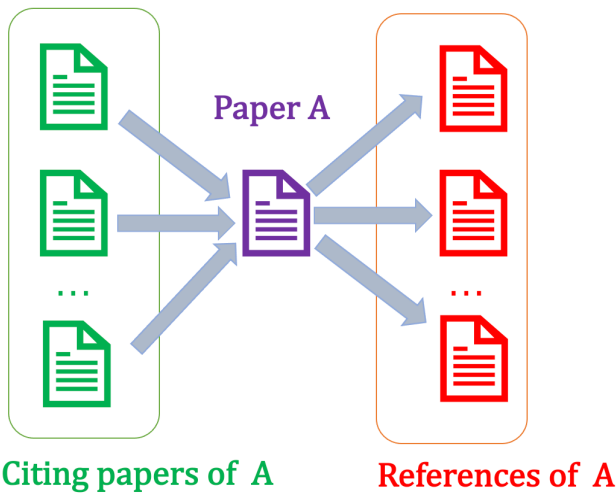

Fig 4. The citation network of a biomedical paper A after its publication. Note that the arrows point from the citing papers to the cited papers (i.e., the references).

Specifically, for the paper dimension, we associated each biomedical paper with twenty-one features, including three citation-related features, four clinical translation-related features, six topic-related features. Meanwhile, we also considered eight features that have previously been confirmed to be highly related to the citation count of academic papers, and we call them "other features". We call them "other features". The details of the twenty-one features are discussed below.

**Citation-related features** of a specific biomedical paper include the total number of citations ($C\_N$), the number of references ($n\_ref$), and the proportion of clinical papers in its references ($p\_clin$). Specifically, the total number of citations ($C\_N$) is the number of citations received by the paper N years after publication. The smaller the value of N ($N \geq 1$), the less citation-related information will be added to the model. This study aims at predicting the clinical citation count of biomedical papers with the minimal citation-related information to early discover biomedical research with high potential to be clinically translated. Particularly, we set N to 2 for training the prediction models, i.e., the citation information two-year after the publication of the paper was used.

The number of references ($n\_ref$) is defined as the number of PubMed papers cited by the paper of interest. We used the PMID pairs to count the number of references, and papers that were not indexed in PubMed were excluded. Meanwhile, we classified papers whose article types are clinical trials or clinical guidelines into clinical papers. The proportion of clinical papers in the references ($p\_clin$) can reflect the research content of the paper: if the proportion is higher, the research could be closer to clinical science and may get more clinical citations (Ke, 2020; Urlings et al., 2021).

**Clinical translation-related features** of a specific biomedical paper are translational location (*tl*), the proportion of animal-related MeSH (*a_score*), the proportion of cell/molecular-related MeSH (*c_score*), and the proportion of human-related MeSH (*h_score*). Specifically, for a specific biomedical paper, its translational location means the relative position of the paper on the translational axis, which is a vector pointing from basic science to clinical science (Ke, 2019; Weber, 2013). The value interval of translational location is [-1,1], and the higher the translational location is, the paper is closer to clinical science and more likely to be clinically cited.

In the Triangle of Biomedicine proposed by Weber (2013), MeSH terms are classified into three categories, i.e., the animal-related (**A**) MeSH, the cell/molecular-related (**C**) MeSH, and the human-related (**H**) MeSH. Weber (2013) argued that the A and C MeSH terms are more about basic science, and the H MeSH terms are closer to clinical science. We can use this classification system to represent the clinical translation-related information of a biomedical paper, as MeSH terms indexed by biomedical experts can well reflect the research content of the paper. As shown in Table 3, we can identify the categories of MeSH terms by their tree numbers, which can be obtained from its official website (https://www.ncbi.nlm.nih.gov/mesh/). We also classify biomedical papers into seven categories (i.e., A, C, H, AC, CH, AH, and ACH) by using the combinations of their MeSH terms.

Table 3. Identifying the categories of MeSH with their tree numbers.

| Category | Beginning of Tree Number | Number of MeSH |
|---|---|---|
| Animal-related (**A**) | B01, excluding B01.050.150.900.649.801.400.112.400.400 | 2,479 |
| Cell/Molecular-related (**C**) | A11, B02, B03, B04, G02.111.570, and G02.49 | 3,625 |
| Human-related (**H**) | B01.050.150.900.649.801.400.112.400.400 or M01 | 332 |

**Topic-related features** of a specific biomedical paper include the number of unique disease entities mentioned in its title and abstract (*n_disease*), the number of unique chemical/drug entities mentioned in its title and abstract (*n_drug*), the number of unique protein/gene entities mentioned in its title and abstract (n_gp), the number of all bio-entities mentioned in its title and abstract (*n_ent*), whether a diagnosis or treatment was mentioned in its abstract or abstract (*is_DT*), and the number of MeSH (*n_mesh*). The number of entities and MeSH terms can be easily counted because they have been well-extracted in the PKG. The value of *is_DT* is binary; we obtained it by counting the MeSH terms indexed. Specifically, if a paper has MeSH terms whose tree numbers start with E (excluding E07 equipment and supplies), then the value of *is_DT* of this paper is "1" (i.e., one or more diagnoses or treatments have been mentioned in the paper); or else, the value of *is_DT* will be "0".

**Other features** of a specific biomedical paper. Inspired by the previous studies on citation prediction (Fortunato et al., 2018; Larivière et al., 2015; Leydesdorff et al., 2018; Zhang et al., 2018), we added eight other features, including whether the research has grants (*is_grant*), the number of authors (*n_authors*), the number of countries of authors (*n_countries*), the readability of abstracts (*readability*), the length of title (*title_length*), the length of abstract (*abs_len*), whether the paper was published in top journals (*is_top_journal*), and article type (*pt*). Note that the readability of the abstract was calculated by the Flesch Reading Ease formula (Farr et al., 1951).

For the reference dimension, we associated a biomedical paper with thirty-five features representing the properties of papers cited by the paper of interest. These features include six citation-related features, twelve clinical translation-related features, and seventeen topic-related features. Finally, for the citing paper dimension, there are also thirty-five features that represent the properties of the papers cited the paper of interest. These features also consist of six citation-related features, twelve clinical translation-related features, and seventeen topic-related features. The structure of the features is shown in Fig. 5. The readers can also find the details of all the ninety-one features in Appendix A.

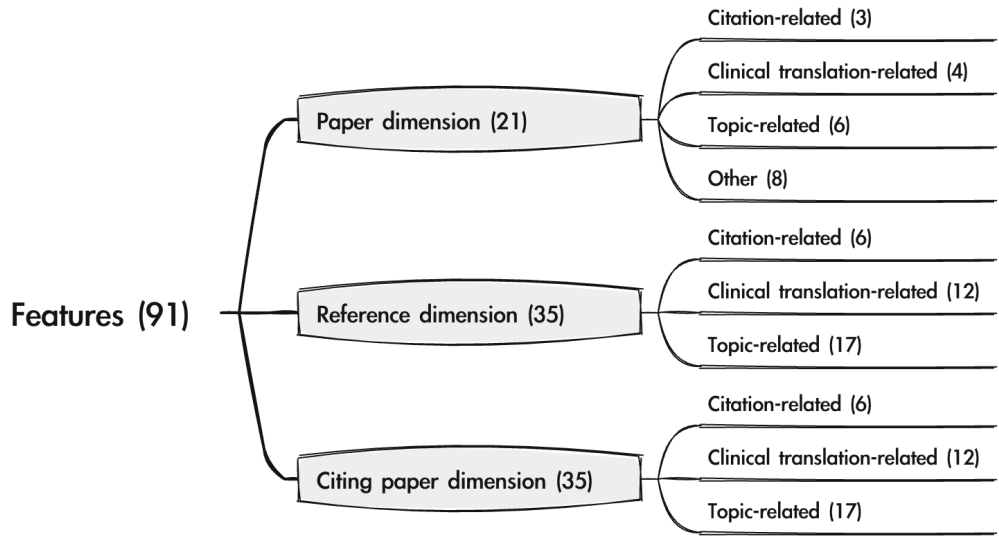

Fig 5. The structure of the ninety-one features of biomedical papers.

To reduce the analysis error and accelerate the convergence speed of the model, for continuous features (e.g., *n_ref* and *n_ent*), we normalized them with the z-score, which is given by:

$$x' = \frac{x - \mu}{\sigma}$$

where $\mu$ means the mean of the values of feature x, $\sigma$ represents the standard deviation of the values of feature x, and $x'$ is the normalized value of feature x. The mean and standard deviation of the normalized features are 0 and 1, respectively. For the binary features, such as *is_DT* and

*is_top_journals*, we used the OneHotEncoder[1], which is the process by which categorical data are converted into numerical data, to normalize them.

### 3.3 The MPNN model for predicting clinical citation counts

3.3.1 The architecture of our MPNN model

We designed a neural network model (see Fig. 6) for predicting the clinical citation count of biomedical papers based on a classic deep neural network architecture, i.e., multilayer perceptron neural network (MPNN). The MPNN has been successfully used for multiple tasks, such as link prediction and entity prediction; it consists of an input layer, an output layer, and one or more hidden layers (Gardner & Dorling, 1998; Tolstikhin et al., 2021). To avoid overfitting in model training, we added dropout layers between every dense layer. The nonlinear transformation in each step makes it possible for the model to learn more information from the paper representations. The MPNN model aimed at predicting the clinical citation count of biomedical papers based on the paper representations. The number of nodes in the input layer is equal to the number of paper features, i.e., ninety-one. The output of our model is the predicted clinical citation count of a biomedical paper; thus, the number of nodes in the output layer is one. Meanwhile, we designed the MPNN model with two hidden layers, and the numbers of the first and second hidden layers are 130 and 80, respectively. The details on how to determine the hidden layers of the MPNN model are discussed in section 3.3.2.

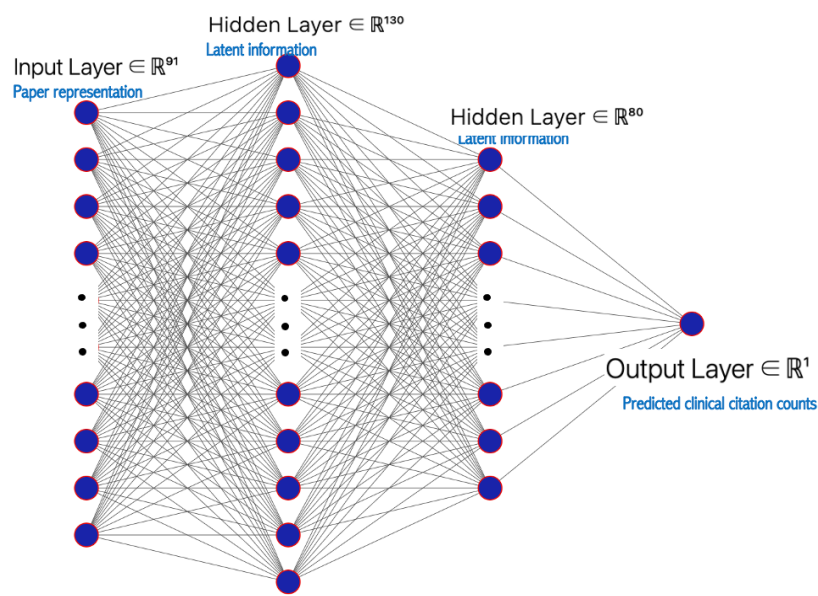

Fig 6. The architecture of the multilayer perceptron neural network (MPNN) model for predicting the clinical citation count of biomedical papers.

3.3.2 Parameters of the MPNN model

The parameters of our model include the size of hidden layers, the activation function in each hidden layer, the learning rate, the optimization method for the loss function, etc. In this study, we used the tenfold cross-validation and grid search to determine appropriate parameters according to the average value of MSE (Ruan et al., 2020). The information on the parameters of the MPNN model is displayed in Table 4.

---

[1] https://scikit-learn.org/stable/modules/generated/sklearn.preprocessing.OneHotEncoder.html

Table 4. Parameters of the MPNN model for predicting clinical citation counts of biomedical papers.

| Parameter | Description | Search range | Value |
| --- | --- | --- | --- |
| The size of hidden layers | The number of hidden layers and the number of nodes in each hidden layer. | / | (130, 80,) |
| Activation | The activation function in each hidden layer. | {"logistic", "tanh", "relu"} | "relu" |
| Solver | The optimization method for loss function. | {"lbfgs", "sgd", "adam"} | "adam" |
| Learning rate (lr) | The update rate of weights. | / | Initial value: 0.01; loss<2.6, lr=0.005; loss<2.4, lr=0.001; loss<2.2, lr = 0.0005; and loss<2.0, lr=0.0001. |
| Dropout rate | The probability that nodes will be dropped with. | [0.0,0.1,0.2,0.3,0.4,0.5,0.6,0.7,0.8,0.9] | 0.3 |

For determining the size of the hidden layer, we used the following steps. First, for a single hidden layer, we set the initial number of nodes in the hidden layer to 10 and trained and tested the model with 10-fold cross-validation and the average MSE. Then, the number of nodes of the hidden layer was gradually increased (10 for each time and up to 200), and the 10-fold cross-validation was used for model training and testing to obtain the optimal number of nodes, with which the MSE of the model reached its minimum. Second, on the premise of fixing the parameters of the previously hidden layers, we added a new hidden layer with the initial number of nodes (10); and we then determined its optimal number of nodes according to the first step. Third, repeat the second step. As shown in Fig. 7, we find that the performance of the MPNN model with only one hidden layer is worse than that of the model with multiple hidden layers. Meanwhile, there is no significant difference between the performance of the MPNN model with two hidden layers and that of the model with three or more hidden layers. Finally, we select the MPNN model with two hidden layers, whose number of nodes are 130 and 80, respectively.

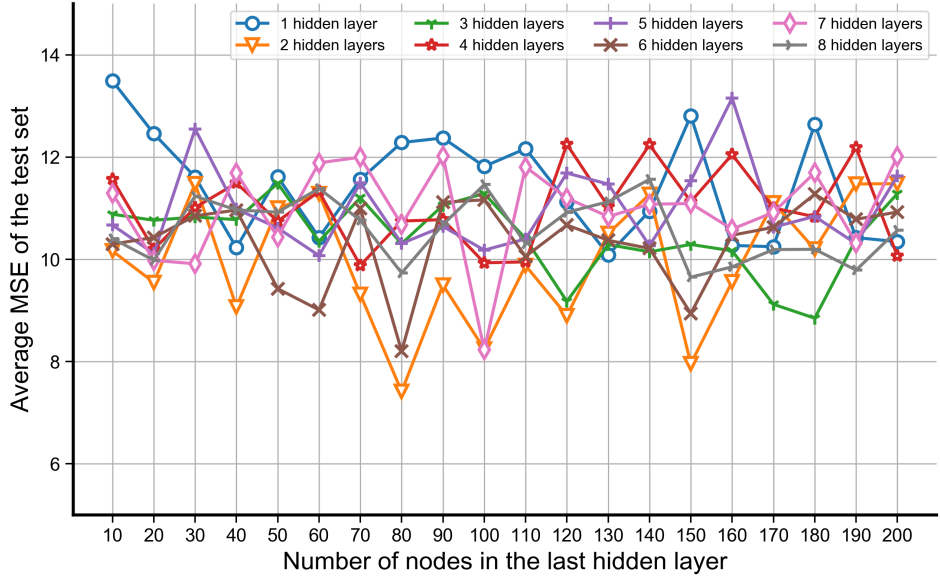

Fig 7. The average MSE of the MPNN model with different numbers of hidden layers and nodes

The learning rate has shown significant effects on the performance of MPNN models in previous studies. The model may converge to a local optimum with a high learning rate and will be time-consuming with a low learning rate. In this study, we used the dynamic learning rate to balance the two ends of the learning rate. Specifically, a higher learning rate (0.01) was used at the initial stage of model training, and it gradually declines as the value of the loss function (i.e., MSE) declines. Eq. (1) displays the loss function of our model:

$$\text{MSE} = \frac{1}{N}\sum_{i=1}^{N}(y_i - y_{predicted})^2 \quad (1)$$

where $y_{predicted}$ is the predicted clinical citation count of paper $i$, and $y_i$ means the observed clinical citation count of paper $i$ in PubMed. Besides, we set the dropout rate as 0.3 to avoid overfitting.

3.3.3 Baseline models and evaluation indicators
(1) Baseline models
In this study, we compared our MPNN model with five other baseline models, including Linear Regression (Persson, 2017), Support Vector Regression (Smola & Schölkopf, 2004), Random Forest Regression (Hutchins et al., 2019b; Smith et al., 2013), K-nearest Neighbors Regression (Song et al., 2017), and eXtreme Gradient Boosting (Chang et al., 2018). These five regression models have been demonstrated to have outstanding performance in previous research (Huang et al., 2022; Hutchins et al., 2019b; Ruan et al., 2020). We implemented these models using the scikit-learn library. Meanwhile, to determine satisfactory parameters of these models, the methods of tenfold cross-validation and grid search were used. The details of the parameters of these five baseline models are shown in Table 5.

(2) Evaluation indicators

In this study, we used the Mean Square Error (MSE), the Mean Absolute Error (MAE), and the R-squared ($R^2$) to evaluate the performance of all the regression models. The definitions of these three indicators are as follows.

**a. Mean Square Error (MSE)** measures the average squared difference between the predicted clinical citation count returned by regression models and the observed values in PubMed. It can be calculated by Eq. (1). The MSE is one of the most used evaluation indicators for a regression model. A smaller value of MSE indicates a better performance of the regression model.

**b. Mean Absolute Error (MAE)** measures the average absolute difference between the predicted clinical citation count returned by a regression model and the observed values in PubMed. It can be calculated by:

$$\text{MAE} = \frac{1}{N} \times \sum_{i=1}^{N} |y_i - y_{predicted}| \qquad (2)$$

compared with MSE, the magnitude of MAE is consistent with the original data, which can represent the actual error of prediction. Similar to MSE, the smaller the MAE is, the better the performance of the regression model is.

**c. R-squared ($R^2$)** is defined as the proportion of the variance for a predicted variable that is explained by features in a regression model. It can be calculated by:

$$R^2 = 1 - \frac{\sum_{i=1}^{N}(y_{predicted} - y_i)^2}{\sum_{i=1}^{N}(y_{mean} - y_i)^2} \qquad (3)$$

where $y_{mean}$ is the mean value of the observed clinical citation count of all samples in PubMed. The interval of the value of $R^2$ is [0,1], and in contrast to MSE and MAE, the higher the value of $R^2$, the better the performance of the regression model. Note that, in rare cases, the value of $R^2$ may be less than zero, indicating that the performance of the prediction model is worse than that of the random benchmark.

Table 5. Parameters of the five baseline models[1].

| Model | Parameter | Description | Search range | Value |
|---|---|---|---|---|
| Linear Regression | fit_intercept | Whether to compute intercept. | {True, False} | True |
|  | n_jobs | Parallel number of jobs. | / | -1 |
|  | copy_X | Whether to copy the training data. | {True, False} | True |
| Support Vector Machine | C | Parameter for regularization. | (1.0, 30) | 25 |
|  | kernel | Category of kernel. | {"linear", "poly", "rbf", "sigmoid", "precomputed"} | "linear" |
| Random Forest | n_estimator | Maximum number of spanning trees. | (10, 2000) | 77 |
|  | max_depth | Maximum depth of trees. | (10, 100) | 23 |
|  | max_features | Number of subsets of features. | {"auto", "sqrt", "log2"} | "sqrt" |
|  | min_sample_leaf | Minimum number of samples required to split nodes. | (1, 30) | 14 |
|  | min_sample_leaf | Minimum number of samples required to determine leaf nodes. | (1, 30) | 6 |
|  | n_jobs | Parallel number of jobs. | / | -1 |
| K-nearest neighbors | n_neighbours | Number of neighbor nodes. | (1, 50) | 7 |
|  | weights | Weight function for prediction. | {"uniform", "distance"} | "distance" |
|  | leaf_size | Size of leaf passing to BallTree or KDTree. | (5, 50) | 18 |
|  | n_jobs | Parallel number of jobs. | / | -1 |
| XGBoost | n_estimators | Maximum number of spanning trees. | (50, 2000) | 750 |
|  | max_depth | Maximum depth of trees. | (1, 100) | 10 |
|  | n_jobs | Parallel number of jobs. | / | -1 |

[1] For parameters that are not listed in the table, we used their default values.

# 4 Prediction result

4.1 The performance of the six regression models

The performance of the six regression models is shown in Table 6, from which we can find that the MPNN model has achieved the best performance in all three indicators (MSE=7.4135, MAE=0.5132, and R-squared=0.7883). The performance of the XGBoost ranks at the second place (MSE = 7.6813, MAE=0.5227 and R-squared=0.7822), which followed by the Random Forest model (MSE = 7.9173, MAE=0.5844 and R-squared=0.7798) and the SVM model (MSE = 8.0652, MAE=0.7611 and R-squared=0.7360). The performance of the KNN model is the worst (MSE=10.377, MAE=0.8961, and R-squared=0.6423).

Table 6. The performance of the six regression models

| Model | # Of Features | MSE | MAE | $R^2$ |
| --- | --- | --- | --- | --- |
| Linear Regression (LR) | 91 | 9.8197 | 0.8510 | 0.7031 |
| Support Vector Machine (SVM) | 91 | 8.0652 | 0.7611 | 0.7360 |
| Random Forest (RF) | 91 | 7.9173 | 0.5844 | 0.7798 |
| K-nearest Neighbors (KNN) | 91 | 10.377 | 0.8961 | 0.6423 |
| eXtreme Gradient Boosting (XGBoost) | 91 | 7.6813 | 0.5227 | 0.7822 |
| Multilayer Perceptron Neural Network (MPNN) | 91 | **7.4135** | **0.5132** | **0.7883** |

Particularly, in the above experiment, we only used the citation information within two years after publication, including the citation-related features in the paper dimension (e.g., *C_N*) and all the features in the citing paper dimension (e.g., *max_n_ref_1*, *max_tl_1*, and *sd_is_DT_1*). We further tested the performance of the MPNN model when the entire citation information was included (i.e., the citation information from publication to the year 2020). The result shows that the performance of the MPNN model was slightly improved (MSE=7.1962, MAE=0.4971, and R-squared=0.7903). Meanwhile, when the citation information was reduced to one-year post-publication, the performance of the MPNN model had a significant drop (the MSE increased from 7.4135 to 8.6713, the MAE increased from 0.5132 to 0.6843, and the R-square reduced from 0.7883 to 0.7274).

4.2 The importance of features

We are also interested in the relative importance of different features for the MPNN model. In this study, we employed the method called "Leave One Feature Out" to calculate the relative importance of different features (Bo et al., 2006; Ruan et al., 2020). Specifically, for a given feature, we first trained and tested the MPNN model without it and then used the absolute difference in the average MSE between the new model and the original model to represent the relative importance of this feature. Table 7 lists the top twenty important features of the MPNN models.

Table 7. The rank of the relative importance of different features (top 20)

| Rank | ALL papers | C papers | CA papers | H papers |
| --- | --- | --- | --- | --- |
| 1 | *sd_n_ref* | *sd_n_ref* | *sd_n_ref* | *sd_n_ref* |
| 2 | *max_n_ref* | *sd_n_ent* | *sd_n_ent* | *max_n_ref* |
| 3 | *sd_n_ent* | *max_h_score* | *max_n_ref* | *max_n_ent* |
| 4 | *max_n_ent* | *mean_h_score* | *max_h_score* | *sd_n_ent* |
| 5 | *max_tl* | *max_n_ref* | *sd_c_score* | *max_c_score* |

| 6 | *C_2* | *mean_n_gp* | *mean_h_score* | *mean_n_drug* |
| --- | --- | --- | --- | --- |
| 7 | *mean_h_score* | *sd_tl* | *mean_n_gp* | *max_tl* |
| 8 | *sd_c_score* | *sd_c_score* | *sd_h_score* | *C_2* |
| 9 | *mean_n_drug* | *sd_h_score* | *sd_tl* | *mean_h_score* |
| 10 | *max_c_score* | *max_n_disease* | *max_n_disease* | *max_n_mesh* |
| 11 | *mean_n_ref* | *mean_tl* | *max_a_score_1* | *mean_is_DT* |
| 12 | *sd_n_drug* | *sd_n_disease* | *mean_tl* | *mean_n_ref* |
| 13 | *sd_h_score* | *max_a_score_1* | *mean_n_ref_1* | *max_a_score* |
| 14 | *sd_tl* | *mean_n_ref* | *max_n_drug* | *sd_n_mesh* |
| 15 | *mean_n_gp* | *max_n_drug* | *sd_n_disease* | *sd_n_gp* |
| 16 | *mean_is_DT* | *mean_n_ref_1* | *max_n_drug_1* | *mean_n_ent* |
| 17 | *max_h_score* | *max_n_drug_1* | *mean_n_disease_1* | *mean_n_mesh* |
| 18 | *mean_n_mesh* | *max_tl_1* | *max_n_gp_1* | *sd_h_score* |
| 19 | *max_n_disease* | *mean_n_drug_1* | *sd_is_DT_1* | *sd_c_score* |
| 20 | *max_a_score* | *max_n_disease_1* | *mean_n_drug_1* | *sd_n_drug* |

In terms of all biomedical papers in the training dataset, we can observe four interesting findings on the importance of features. First, the importance of the standard deviation and the maximum of the references of all references (i.e., *sd_n_ref* and *max_n_ref*) rank the first and second place; the value of the importance of these two features are 1.2762 and 1.0274, respectively. The third and fourth important features are the standard deviation and the maximum of the number of bio-entities mentioned in all references, i.e., sd_n_ent (0.8364) and *max_n_ent* (0.6842). The importance of the total citation counts two years post-publication (0.3883) ranks the fifth place. Second, 83 out of the 91 features have an importance that is not zero.

Third, in terms of the feature dimension, the most important feature dimension is the reference dimension: 32 (91.4%) features in the reference dimension have non-zero importance, and 19 (54.3%) features in the reference dimension rank in the top 20. The second most important feature dimension is the paper dimension: 19 (90.5%) features in the paper dimension have non-zero importance, one feature (*C_2*) in the paper dimension ranks in the top 20 (i.e., the 6th place), and three features in the paper dimension rank in the top 30. The least important feature dimension is the citing paper dimension; although 32 (91.4%) features in the citing paper dimension have non-zero importance, only one feature ranks in the top 30, i.e., the feature ranking in the 30th place "*sd_n_ent_1*".

Fourth, in terms of the feature categories, the citation-related information (such as *sd_n_ref max_n_ref*, and *C_2*) and the clinical topic-related features (e.g., *max_n_ent* and *sd_n_ent*), who rank the first four places, are more important than the clinical translation-related features (e.g., *max_tl* and *mean_h_score*). The importance of other features such as *abs_len* and *is_grant* is the lowest. Besides, the biomedical entity-related features (e.g., *sd_n_ent* and *max_n_ent*) are overall more important than the MeSH term-related features (e.g., *max_n_mesh* and *mean_n_mesh*).

We also analyzed the importance of features for the MPNN models based on different categories of biomedical paper sets, including C papers, CA papers, and H papers. As shown in Table 7, the standard deviation of the number of references of all references (i.e., *sd_n_ref*) is the most important feature for prediction models based on all categories of biomedical paper sets. Meanwhile, in terms of feature dimensions, features in the reference dimension and paper dimension are more important than those in the citing paper dimension. We also find that features in the citing paper dimension are more important for C and CA papers rather than all and H papers. In addition, in terms of feature categories, clinical translation-related features are more important for C and CA papers, and citation-related and topic-related features are more important for H papers.

## 5 Discussion and conclusion

This study proposed a multilayer perceptron neural network (MPNN) model with two hidden layers to predict the clinical citation count of biomedical papers published from 1985 to 2005 in PubMed. Accordingly, we extracted citation-related, clinical translation-related, topic-related features of biomedical papers from three dimensions, including the paper dimension, the reference dimension, and the citing paper dimension. We also considered the features that have previously been demonstrated to have influence on the citation count of academic papers. The PubMed Knowledge Graph (PKG) was employed as the data source in this study because it contains more than 30 million biomedical papers with well-processed information, such as bio-entities, disambiguated author names, and citation relationships between PubMed papers (Xu et al., 2020).

One of the contributions of this study is the improvement of the performance of the clinical citation count prediction model by using the multilayer perceptron neural network. The results of the experiment demonstrated that the proposed MPNN model performed significantly better than the other five baseline models, including eXtreme Gradient Boosting, random forest, support vector machine, linear regression, and K-nearest neighbors.

The other contribution of this study is that we found the most important features for clinical citation count prediction is the features in the reference dimension, which should not have been ignored in the previous studies. Moreover, the results of this study showed that the citation-related and topic-related features are more important for clinical citation count prediction than the clinical-translation-related ones. However, by further analyzing MPNN models based on different categories of biomedical papers, we found that the clinical translation-related features are more important for the prediction of clinical citation count of basic papers (i.e., C and CA papers) rather than papers closer to clinical science (i.e., H papers). This could be interpreted as those basic papers need obvious clinical translation-related features to be found and cited by the authors of clinical papers, but papers closer to clinical science (H papers) don't need them. In addition, we also demonstrated "other features" that were helpful in the prediction of the citation count of academic papers, are not important for the prediction of the clinical citation count of biomedical papers. This indicates that there is essential difference between the academic citations and the clinical citations of biomedical papers.

We analyzed the distribution of the clinical citation count of biomedical papers by using exploratory data analysis. The results showed that the clinical citation count of biomedical papers tend to be stable fifteen years post-publication, which demonstrated that it is feasible for us to select biomedical papers published during 1985-2005. Meanwhile, highly similar distribution patterns were found from both the total citations and clinical citations, which indicated that the clinical citation count prediction of biomedical could be possible, just as that of the previous citation counts. We believe that these findings will be useful for the follow-up research on the clinical citation count of biomedical papers.

There are several implications of this study. Methodologically, this paper predicted the clinical citation count of biomedical papers using the multilayer perceptron neural network and achieved good performance, with citation information only two years post-publication. This could be useful for the policymakers, and the pharmaceutical companies to early assess the translational progress of biomedical research and to monitor the biomedical research with a high potential to be clinically translated in real-time. Meanwhile, this study demonstrated that the features in the reference dimension are the most important for the clinical citation count prediction of biomedical papers, but they were rarely studied in previous research (Hutchins et al., 2019b). Therefore, our study provides a novel dimension for the research community and could be applied to other related research tasks, such as the research assessment for translational programs. Besides, the findings in this study could be useful for biomedical authors, especially those in basic science, to get more attention from clinical research.

The study also has several limitations. First, we only included ninety-one features in this study for predicting the clinical citation count of biomedical papers. Several other factors may also be considered, such as the image-related factors, the length of the whole paper, and full-text embeddings. Second, the landscape of collaborations between clinical science and basic science may also affect the clinical citations of biomedical papers. Future work should include all these factors for the model training. Several knowledge representation methods such as word2vec, BERT or graph neural network could also be used for paper representation in the step of feature extraction. Third, the causal relationships between the features and the clinical citation count of biomedical papers were not determined because of the "black box" nature of neural networks. Eventually, in our future work, we also intend to explore the underlying relationships between factors in different dimensions and the clinical citation count of biomedical papers.

## Acknowledgment
This work was supported by the National Natural Science Foundation of China (grant no. 72204090). This work was also supported by "the Fundamental Research Funds for the Central Universities" (grant no. CCNU22XJ025). The computation is completed in the HPC Platform of Huazhong University of Science and Technology. The computation is completed in the HPC Platform of Huazhong University of Science and Technology.

# Appendix A. The details of the ninety-one features of biomedical papers

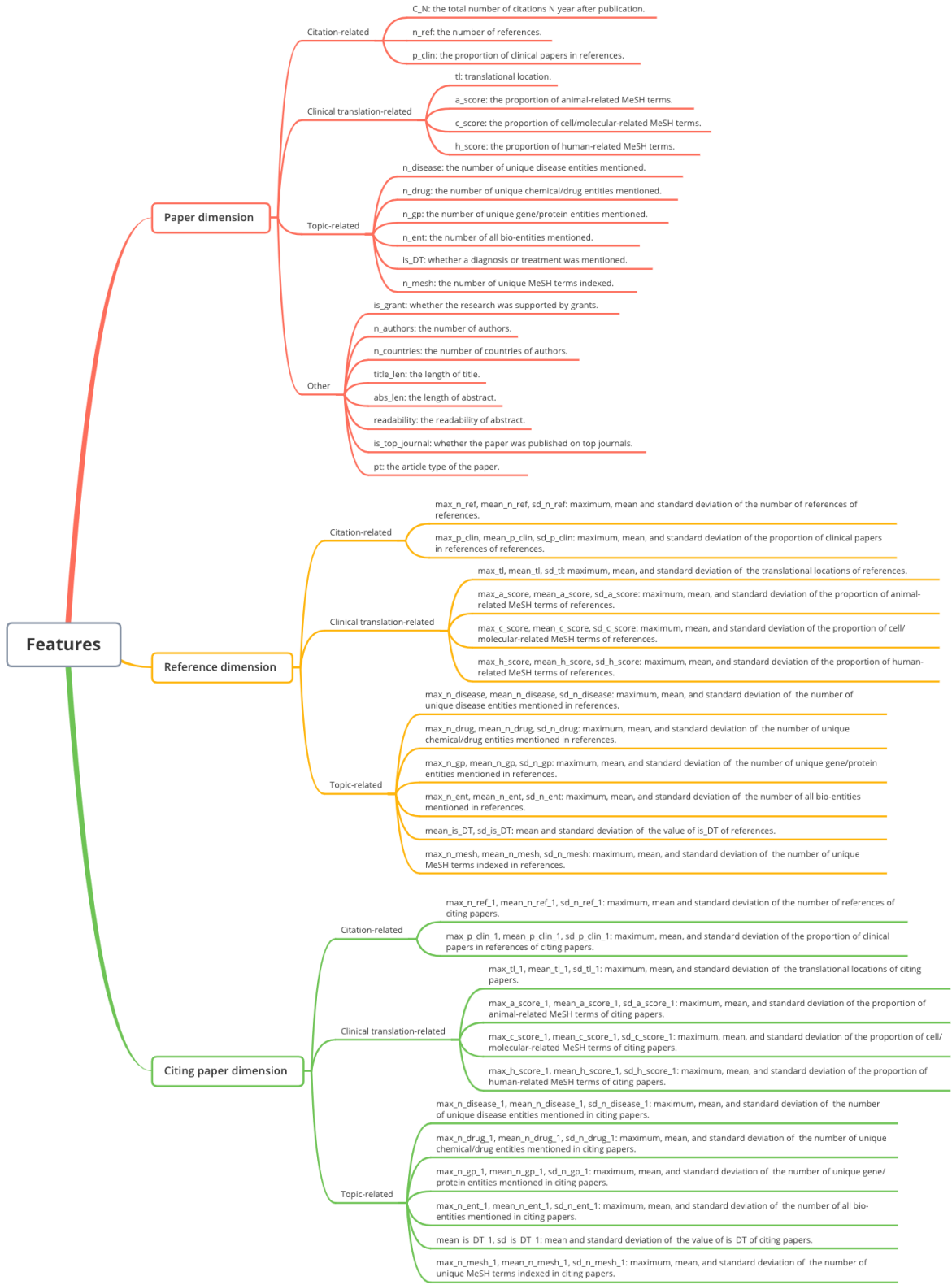